\UseRawInputEncoding
\documentclass[times,twocolumn,final,authoryear]{elsarticle}

\usepackage{ycviu}
\usepackage{framed,multirow}

\usepackage{amssymb}
\usepackage{latexsym}
\usepackage{comment}
\usepackage{hyperref}
\usepackage{inputenc}

\usepackage{url}
\usepackage{xcolor}
\definecolor{newcolor}{rgb}{.8,.349,.1}

\journal{Computer Vision and Image Understanding}

\begin{document}

\thispagestyle{empty}

\clearpage

\ifpreprint
  \setcounter{page}{1}
\else
  \setcounter{page}{1}
\fi

\begin{frontmatter}

\title{Periocular Biometrics and its Relevance to Partially Masked Faces: A Survey}

\author[1]{Renu \snm{Sharma}\corref{cor1}} 
\cortext[cor1]{Corresponding author: }
\ead{sharma90@cse.msu.edu}
\author[1]{Arun \snm{Ross}}

\address[1]{Department of Computer Science and Engineering, Michigan State University, East Lansing, MI 48824, USA}

\received{1 May 2013}
\finalform{10 May 2013}
\accepted{13 May 2013}
\availableonline{15 May 2013}
\communicated{S. Sarkar}

\begin{abstract}
The performance of face recognition systems can be negatively impacted in the presence of masks and other types of facial coverings that have become prevalent due to the COVID-19 pandemic. In such cases, the periocular region of the human face becomes an important biometric cue. In this article, we present a detailed review of periocular biometrics. We first examine the various face and periocular techniques specially designed to recognize humans wearing a face mask. Then, we review different aspects of periocular biometrics: (a) the anatomical cues present in the periocular region useful for recognition, (b) the various feature extraction and matching techniques developed, (c) recognition across different spectra, (d) fusion with other biometric modalities (face or iris), (e) recognition on mobile devices, (f) its usefulness in other applications, (g) periocular datasets, and (h) competitions organized for evaluating the efficacy of this biometric modality. Finally, we discuss various challenges and future directions in the field of periocular biometrics. \end{abstract}


\end{frontmatter}


\section{Introduction}
\label{sec:Periocular-Introduction}
Biometrics is the automated or semi-automated recognition of individuals based on their physical (face, iris), behavioral (signature, gait), or psychophysiological (ECG, EEG) traits \citep{Jain2011, Ross2019}. The COVID-19 pandemic has ushered in a number of considerations for biometric systems. For example, in the context of fingerprint recognition, researchers are now investing more effort in designing contactless fingerprint systems \citep{Yin2020,Kumar2019b}. Similarly, the prevalent use of face masks and social distancing protocols has refocused attention on occluded face recognition and, inevitably, ocular biometrics. 
The ocular region refers to the anatomical structures related to the eyes, and biometric cues in this region include pupil, iris, sclera, conjunctival vasculature, periocular region, retina, and oculomotor plant (comprising eye globe, muscles, and the neural control signals).

\begin{figure}[h]
\centering
\includegraphics[width=\columnwidth]{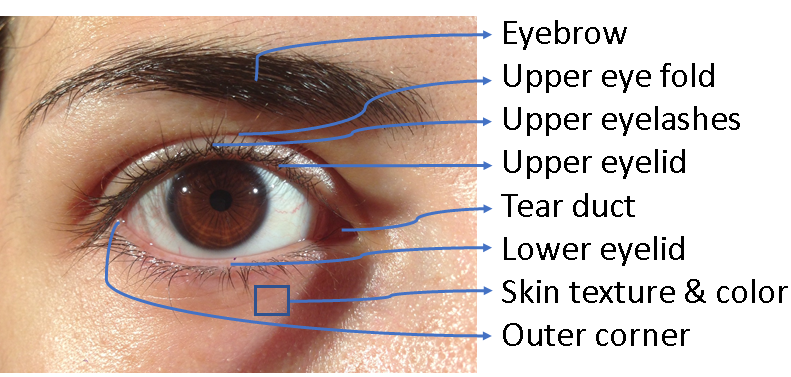}
\caption{Various components of the periocular region when viewed in the visible spectrum.}
\label{fig:periocular-components}
\end{figure}

The term ``periocular" has been used to refer to the region surrounding an eye consisting of eyelids, eyelashes, eye-folds, eyebrows, tear duct, inner and outer corner of an eye, eye shape, and skin texture (Figure \ref{fig:periocular-components}). While many articles in the biometric literature include the sclera, iris, and pupil in the context of periocular recognition \citep{Park2009, Miller2010b, DeMarsico2017, Smereka2017, Luz2018}, others have excluded these regions \citep{Woodard2010a, Woodard2010b, Park2011, Proenca2018}. The periocular region may be biocular (the periocular regions of both eyes are considered to be a single unit) \citep{Jillela2012, JuefeiXu2012, Proenca2014b}, monocular (either left or right periocular) \citep{Park2009, Park2011}, or a fusion of the two monocular regions (combination of left and right periocular regions) \citep{Bharadwaj2010,Woodard2010a,Boddeti2011}. Earlier work \citep{Park2009, Park2011} on periocular biometrics studied its feasibility as a standalone biometric trait. Other researchers \citep{Woodard2010b, Park2011, JuefeiXu2012} established its relevance by comparing it with the face and iris modalities. In some non-ideal conditions, the periocular region even shows higher performance than face \citep{Miller2010b, Park2011, JuefeiXu2012} and iris \citep{Boddeti2011} modalities. Hollingsworth et al. ascertained its usefulness as a biometric trait by conducting human analysis on near-infrared \citep{Hollingsworth2010, Hollingsworth2011, Hollingsworth2012} and visible \citep{Hollingsworth2012} spectrum images.

\begin{figure*}[h]
\includegraphics[width=\textwidth]{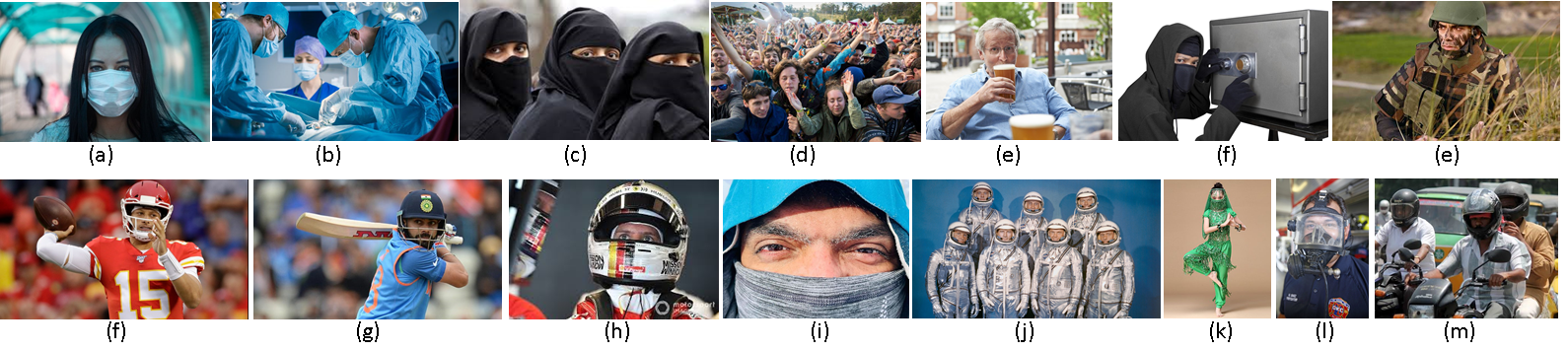}
\caption{Various scenarios where the periocular region has increased significance: (a) girl with a mask during the covid pandemic, (b) doctors and nurses in the surgical room, (c) women in niqab, (d) partial faces in the crowd, (e) occluded face while drinking, (f) robber with face covering, (e) military personnel with face paint, (f) football player wearing a helmet, (g) cricket player wearing a helmet, (h) F1 race player in a helmet, (i) face-covering in cold weather, (j) astronauts in suit, (k) dancer with a face veil, (l) firefighter in uniform, and (m) people in motorbike helmets.}
\label{fig:periocular-applications}
\end{figure*}

Periocular recognition has numerous applications that go beyond the current pandemic (Figure \ref{fig:periocular-applications}). This includes (a) operating theaters where physicians wear surgical masks; (b) occupations where people wear helmets that occlude faces (e.g., military, astronauts, firefighters, bomb diffusion squads); (c) sporting events requiring a helmet (cricket, football, car racing); (d) use of veils to cover one's face due to cultural or religious purposes ; and (e) robbers masking their faces to avoid being recognized. 
Periocular biometrics has various advantages, which further motivate its usage:

\begin{enumerate}
\item  Periocular modality can be acquired using the sensors that capture face and iris modalities. So, there is typically no additional imaging requirement.

\item Compared to iris or other ocular traits (e.g., retina or conjunctival vasculature), periocular images can be captured in a relatively non-invasive, less constrained, and non-cooperative environment. They are also less prone to occlusions due to eyelids, eyeglasses, or deviated gaze. In contrast to the face modality, the periocular region (which is, of course, a part of the face) is relatively more stable as it is less affected by variations in pose, aging, expression, plastic surgery, and gender transformation. It is also seldom occluded when face images are captured in close quarters (e.g., selfies) or in the presence of scarves, masks, or helmets.

\item The periocular modality can complement the information provided by the iris and face modalities. So, it can be combined with the iris \citep{Santos2012, Tan2012b, Raghavendra2013, AlonsoFernandez2015a} and face \citep{Jillela2012, Mahalingam2014} modalities to increase the performance of the biometric system without any modification to the  acquisition setup.

\item It can also be used for other tasks such as presentation attack detection \citep{Hoffman2019}, and soft-biometrics extraction \citep{Merkow2010, Lyle2012, Rattani2017d, AlonsoFernandez2018}.

\item It can help with cross-spectral iris recognition \citep{Santos2015} as iris images captured in different spectra depict different features, while periocular features (shape of the eye, eyelashes, eyebrows) are relatively stable. It also facilitates cross-modal (face-iris) matching \citep{Jillela2014} as it is the common region present in both the modalities.

\end{enumerate}

\begin{figure*}[h]
\includegraphics[width=\textwidth]{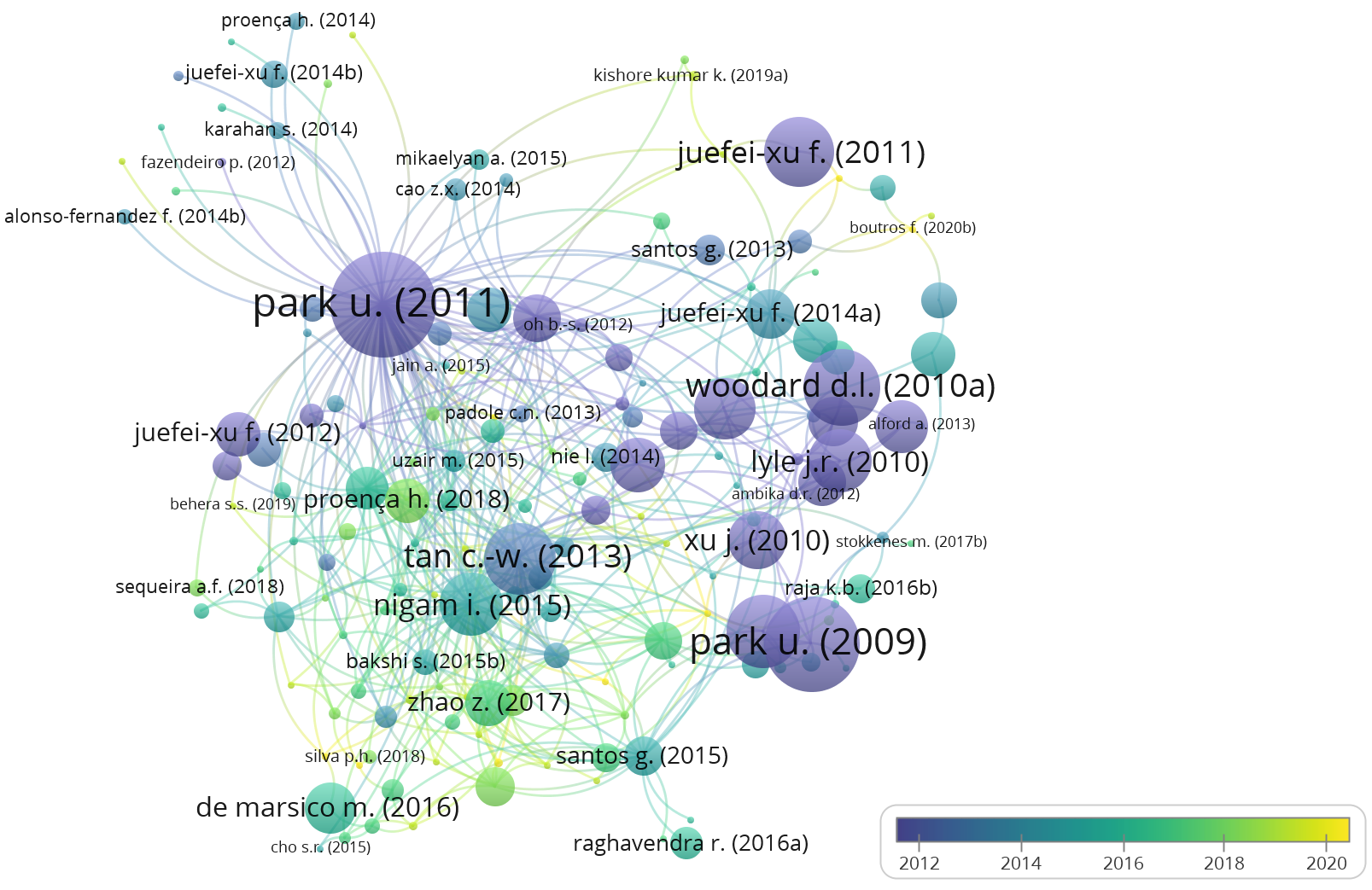}
\caption{Network visualization of research articles on periocular biometrics. The size of the node represents the number of citations, and its color represents the year of publication. Figure generated using VOSviewer software.}
\label{fig:periocular-survey}
\end{figure*}

In the literature, there are previous surveys that focused on periocular biometrics \citep{Santos2013, Nigam2015, AlonsoFernandez2016b, Rattani2017a, Badejo2019,Behera2019,Kumari2019}. \cite{Santos2013} summarized the significant papers on periocular recognition before 2013. The authors in \citep{Nigam2015, Rattani2017a} discussed the research advances of various ocular biometric traits such as iris, periocular, retina, conjunctival vasculature and eye movement, and their fusion with other modalities. \cite{AlonsoFernandez2016b} described periocular recognition methodologies in terms of pre-processing, feature extraction, fusion, soft-biometric extraction, and other applications.  
\cite{Behera2019} focused on cross-spectral periocular recognition. \cite{Zanlorensi2021} provided detailed information on periocular and iris datasets, and competitions based on some of these datasets. A recent survey on periocular biometrics is in \citep{Kumari2019}. Figure \ref{fig:periocular-survey} shows a visualization of various research work on periocular biometrics. The main contributions of this paper lie in the detailed categorization of periocular techniques. We also describe periocular techniques specifically useful for human recognition in the COVID-19 pandemic. 

In this paper, we present a comprehensive review of periocular biometrics. We discuss different categorizations based on (a) different anatomical cues utilized for recognition, (b) feature extraction or matching methodologies, (c) different spectral input images, and (d) fusion with different modalities. We then discuss periocular recognition techniques for mobile devices, in other applications (e.g., soft biometrics, iris presentation detection) and for recognition in special circumstances (cross-modal, gender transformation, long-distance). We also describe various periocular datasets and competitions held. Considering the COVID-19 pandemic situation, we also provide a brief review of recent face and periocular techniques specifically designed to recognize humans wearing a face mask. 

The rest of the paper is organized as follows: Section \ref{sec:Periocular-COVID19} describes various face and periocular techniques specially applied on the masked faces for human identification, Section \ref{sec:Periocular-OtherCues} categorizes periocular techniques based on anatomical cues utilize for recognition, Section \ref{sec:Periocular-Methodologies} describes various periocular features extraction and matching techniques, Section \ref{sec:Periocular-Spectrums} categorizes techniques based on input images of different spectra, Section \ref{sec:Periocular-Fusion} discusses fusion techniques with other biometric modalities, Section \ref{sec:Periocular-MobileDevices} provides details of periocular authentication on mobile devices, Section \ref{sec:Periocular-OtherApplications} describes periocular recognition in specific scenarios and other applications, Section \ref{sec:Periocular-Datasets} details periocular datasets and competitions, Section \ref{sec:Periocular-Challenges} focuses on various challenges and future directions, and Section \ref{sec:Periocular-Conclusion} concludes the paper.

\section{Periocular in COVID-19 Pandemic}
\label{sec:Periocular-COVID19}
Periocular recognition has gained relevance during the COVID-19 pandemic as some reports have documented a drop in performance of existing face recognition methods in the presence of facial masks \citep{Damer2020, Ngan2020a, Ngan2020b}. The tests conducted by NIST applied digitally tailored masks to face images for evaluation (6.2 million images from 1 million people). The first report \citep{Ngan2020a} presented the performance of 89 algorithms that were submitted to NIST before the COVID-19 pandemic on the masked images. All 89 face recognition algorithms showed an increase in False Non-Match Rate (FNMR) by about 5\% - 50\% at a 0.001\% False Match rate (FMR) –- higher than NIST’s prior study on unmasked images. The second report \citep{Ngan2020b} published the performance of 65 new algorithms submitted to NIST after mid-March 2020 along with their previous submissions (cumulative results for 152 algorithms). The new algorithms include masked images during the enrollment stage. However, the report showed increased FNMR (5\% - 40\%) for all the newly submitted algorithms, though the new algorithms showed improved accuracy compared to the pre-pandemic algorithms. An earlier work by \cite{Park2011} also showed a drop in rank-one accuracy of a commercial face recognition software from 99.77\% (full-face images) to 39.55\% when the lower region was occluded. 

In an era of masked faces necessitated by the pandemic, periocular information can be helpful for human recognition in two ways, either by generating a full face from the periocular region or by matching using only the periocular region. Juefei-Xu et al. \citep{JuefeiXu2014b, JuefeiXu2016} hallucinated the entire face from the periocular region using dictionary learning algorithms. \cite{UdDin2020} detected the masked region from a face image and then performed image completion on the masked region. They used a GAN-based network for image completion, which consists of two discriminators; one learns the global structure of the face and the other focuses on learning the missing region. \cite{Li2020} also performed face completion to recover the content under the mask through the de-occlusion distillation framework. \cite{Hoang2020} emphasized the use of eyebrows for human identification. \cite{Wang2020} released three datasets for face and periocular evaluation on masked images: Masked Face Detection Dataset (MFDD), Real-world Masked Face Recognition Dataset (RMFRD), and Simulated Masked Face Recognition Dataset (SMFRD). \cite{Anwar2020} presented an open-source tool, MaskTheFace, to create masked faces images. Moreover, research work on face detection in the presence of masks \citep{Opitz2016, Ge2017}, face mask detection \citep{Chowdary2020, Qin2020, Loey2021} and face recognition under occlusion \citep{Song2019, Ding2020, Geng2020, Boutros2021a,Damer2021, Hariri2021,Montero2021} would be helpful in human identification on masked face images. Various competitions are also conducted to benchmark face recognition techniques on masked faces \citep{Deng2021, Boutros2021b, Zhu2021}.

\section{Anatomical Cues in the Periocular Region}
\label{sec:Periocular-OtherCues}
\cite{Woodard2010b} classified periocular anatomical cues into two levels: level-one cues comprise eyelids, eye folds, eyelashes, eyebrows, and eye corners, while level-two includes skin texture, fine wrinkles, color, and skin pores. Specifically, level-one cues represent geometric nature of the periocular region, while level-two embodies textural and color attributes. The authors in \citep{Hollingsworth2010, Hollingsworth2011, Park2011, Oh2012a} studied the significance of various periocular components in recognizing individuals. Earlier work on face recognition \citep{Sadr2003} suggested the eyebrows to be the most salient and stable feature of the face. Hollingsworth et al. \citep{Hollingsworth2010, Hollingsworth2011} conducted a human analysis to identify the discriminative cues on near-infrared (NIR) images and found that eyelashes, tear ducts, shape of the eye, and eyelids are the most frequently used cues in verifying the two images of a person. The studies in \citep{Park2011, Oh2012a} utilized automatic feature descriptors to determine important regions on visible (VIS) spectrum images and concluded that eyebrows, iris, and sclera are the most significant cues for periocular performance. In a subsequent work \citep{Hollingsworth2012}, the authors applied both human and machine approaches to identify discriminative regions on both NIR as well as VIS periocular images. They observed that humans and computers both focus on the same periocular cues for identification: in VIS images, blood vessels, skin region, and eye shape are more salient, whereas in NIR images, eyelashes, tear ducts, and eye shape are more promising. Other authors \citep{Smereka2013, AlonsoFernandez2014b, Smereka2017} also drew similar conclusions on the relevance of periocular cues in VIS and NIR images. In summary, level-one cues are more useful for NIR images, whereas level-two cues aid in VIS images.

Researchers have also analyzed the utility of periocular cues as a standalone biometrics, for instance, using only the eyebrows \citep{Dong2011, Le2014, Hoang2020}, or periocular skin \citep{Miller2010a}, or eyelids \citep{Proenca2014a}. Details of the other ocular biometrics traits closely related to periocular can be found in the following studies: iris \citep{Bowyer2008, Bowyer2013}, sclera \citep{Zhou2012, Das2013}, conjunctiva vasculature \citep{Derakhshani2007, Crihalmeanu2012}, eye movements \citep{Rigas2012a, Holland2013, Sun2014}, occulomotor plant characteristics \citep{Komogortsev2010}, and gaze analysis \citep{Cantoni2015}. The description of these ocular traits is out of the scope of this paper.

\section{Methodologies used for Periocular Recognition}
\label{sec:Periocular-Methodologies}
A typical periocular recognition system consists of the following steps: acquisition, pre-processing of the acquired image, localization of region-of-interest (ROI), feature extraction, post-processing of extracted features, and matching of two feature sets. In the acquisition step, the periocular image is captured using a sensor or camera. We provide details of various sensors used to capture periocular images along with their datasets in Section \ref{sec:Periocular-Datasets}. The pre-processing step aims to enhance the visual quality of an image. Commonly, pre-processing techniques are applied to normalize illumination variations, such as anisotropic diffusion \citep{JuefeiXu2012} and Multiscale Retinex (MSR) \citep{JuefeiXu2014b, Nie2014}. \cite{Karahan2014} applied histogram equalization for contrast enhancement. \cite{JuefeiXu2011b} performed pre-processing schemes for pose correction, illumination, and periocular region normalization. \cite{Proenca2014c} investigated an elastic graph matching (EGM) algorithm to handle non‐linear distortions in the periocular region due to facial expressions.

The localization step extracts the periocular region from the acquired or pre-processed image. As the definition of the periocular region has not yet been standardized, the ROI used for periocular recognition varies across the literature. The authors in \citep{Tan2012b,Park2009, Mahalingam2014} considered the iris center as a reference point to determine the periocular rectangular region. The authors in \citep{Padole2012, Nie2014} used the geometric mean of eye corners to localize the ROI since the iris center is affected by gaze, pose, and occlusion. \cite{Bakshi2013} localized the periocular region based on the anthropometry of the human face. \cite{Park2011} studied the effect of including eyebrows in ROI on the recognition performance by performing both manual localization (based on the centers of the eyes) and automatic localization (based on the anthropometry of the human face). \cite{Algashaam2017b} analyzed the influence of varying periocular window sizes on periocular recognition performance. \cite{Kumari2021b} proposed an approach to extract optimum size periocular ROIs of two different shapes (polygon and rectangular) by using five reference points (inner and outer canthus points, two end points and the midpoint of eyebrow). \cite{Proenca2014b} described an integrated algorithm for labeling seven components of the periocular region in a single-shot: iris, sclera, eyelashes, eyebrows, hair, skin, and glasses. Deep learning techniques have also been used to detect the periocular region, such as ROI-based object detectors \citep{Reddy2018a} and supervised semantic mask generators \citep{Zhao2018}. \cite{Reddy2020} proposed spatial transformer network (STN), which is trained in conjunction with the feature extraction model to detect the ROI.

The feature extraction step involves the extraction of discriminative and robust features from the localized periocular region. \cite{AlonsoFernandez2016b} categorized feature extraction techniques into global and local approaches. We group deep learning-based approaches separately. Table \ref{table:periocular-features} lists the various feature extraction techniques corresponding to these categories, along with research papers utilizing these techniques. The description of all three approaches are provided below.

1. \textbf{Global Feature Approaches:} The global feature extraction approaches consider the entire periocular ROI as a single unit and extract features based on texture, color, or shape. Texture in a digital image refers to the repeated spatial arrangement of the image pixels. Commonly used techniques to capture the textural features from the periocular region are Local Binary Patterns (LBP) and its variants \citep{Park2009,Adams2010a,Bharadwaj2010,JuefeiXu2010, Miller2010a,Xu2010,JuefeiXu2012,Oh2012a,Padole2012,Santos2012,Uzair2013, Cao2014a, Mahalingam2014, Nie2014, Sharma2014, Santos2015}, Histogram of Oriented Gradients (HOG) \citep{Park2009, Park2011}, Gabor filters \citep{JuefeiXu2010,AlonsoFernandez2012, Joshi2014, Cao2014a,AlonsoFernandez2015a}, and Binarized Statistical Image Features (BSIF) \citep{Raghavendra2013, Raja2014a}. The LBP descriptor computes the binary patterns around each pixel by comparing the pixel value with its neighborhood. The binary patterns are then quantized into histograms, which on concatenation form a feature vector. In the HOG descriptor, gradient orientation and magnitude around each pixel are binned into histograms and histograms are then concatenated to form a feature vector. The Gabor filters extract features by applying textural filters of different frequencies and orientations on an image. The BSIF descriptor convolves the image with a set of filters learned from natural images, and then the responses are binarized. Other texture-based features include Bayesian Graphical Models (BGM) \citep{Boddeti2011}, Probabilistic Deformation Models (PDM) \citep{Ross2012,Smereka2013}, Discrete Cosine Transform (DCT) \citep{JuefeiXu2010}, Discrete Wavelet Transform (DWT) \citep{JuefeiXu2010,Joshi2014}, Force Field Transform (FFT) \citep{JuefeiXu2010}, GIST perceptual descriptors \citep{Bharadwaj2010}, Joint Dictionary-based Sparse Representation (JDSR) \citep{Raghavendra2013,Jillela2014,Moreno2016}, Laws masks \citep{JuefeiXu2010}, Leung-Mallik filters (LMF) \citep{Tan2012b}, Laplacian of Gaussian (LoG) \citep{JuefeiXu2010}, Correlation-based methods \citep{Boddeti2011, JuefeiXu2012, Ross2012, Jillela2014}, Phase Intensive Global Pattern (PIGP) \citep{Smereka2013, Bakshi2014}, Structured Random Projections (SRP) \citep{Oh2014}, Walsh masks \citep{JuefeiXu2010},  Higher Order Spectral (HOS) \citep{Algashaam2017b}, Gaussian Markov random field \citep{Smereka2015}, and Maximum Response (MR) \citep{Raghavendra2016a}. 

The color features of the periocular region correspond to the wavelengths of light reflected from its constituent parts. \cite{Woodard2010b} utilized the color features by applying histogram equalization on the luminance channel and then calculating the color histogram on the spatially salient patches of the image. \cite{Lyle2012} also extracted color features using local color histograms. \cite{Moreno2016} defined color components using linear and nonlinear color spaces such as red-green-blue (RGB), chromaticity-brightness (CB), and hue-saturation-value (HSV) and then applied a re-weighted elastic net (REN) model. The authors in \citep{Woodard2010b, Moreno2016} utilized both textural and color features from the periocular recognition. Regarding shape features, the work in \citep{Dong2011, Le2014} utilized eyebrow shape-based features, while \cite{Proenca2014a} extracted eyelid shape features. \cite{Ambika2016} employed Laplace–Beltrami operator to extract periocular shape characteristics. All aforementioned techniques use 2D image data of the periocular region. \cite{Chen2015} combined 3D shape features extracted using the iterative closest point (ICP) method and fused them with 2D LBP textural features at the score-level. One of the major advantages of using global feature approaches is that they generate feature vectors of fixed-length, and matching of fixed-length vectors is computationally effective. However, global feature approaches are more susceptible to image variations, such as occlusions or geometric transformations.

2. \textbf{Local Feature Approaches:} The local feature extraction approaches first detect salient or key points from the ROI and then extract features from their local neighborhood to create a feature descriptor. Commonly used local feature approaches are Scale Invariant Feature Transformation (SIFT) \citep{Xu2010,Park2011,Padole2012,Ross2012, Santos2012,Smereka2013,AlonsoFernandez2014b} and Speeded-up Robust Features (SURF) \citep{JuefeiXu2010, Xu2010,Raja2015a}. The SIFT feature extractor defines key locations as extrema points on the difference of Gaussians (DoG) images obtained from a series of smoothed and rescaled images. Feature descriptor is then formed by concatenating orientation histograms defined around each key point. On the other hand, SURF detects key points by utilizing the Hessian blob detector, and the key points are then described using Haar wavelet features. SURF utilizes integral images to speed up the computation. Other local feature descriptors are Binary Robust Invariant Scalable Keypoints (BRISK) \citep{Mikaelyan2014}, Oriented FAST and Rotated BRIEF (ORB) \citep{Mikaelyan2014}, Phase Intensive Local Pattern (PILP) \citep{Bakshi2015}, Symmetry Assessment by Feature Expansion (SAFE) \citep{Mikaelyan2014,AlonsoFernandez2015a}, and Dense SIFT \citep{Ahuja2016a}. Since the number of detected key points varies among images, the resulting feature vectors also vary in length, making the process computationally expensive in some cases. However, local feature approaches are more robust to occlusions, illumination variations, and geometric transformations compared to global feature approaches.

3. \textbf{Deep Learning Approaches:} With the success of deep learning in computer vision and biometrics, this approach has also been applied to periocular recognition. Earlier work \citep{Nie2014} based on learning approaches introduced an unsupervised convolutional version of Restricted Boltzman Machines (CRBM) for periocular recognition. Raja et al. \citep{Raja2016a, Raja2020} extracted features from Deep Sparse Filters using transfer learning methodology and input them into a dictionary-based approach for classification. On the other hand, \cite{Raghavendra2016a} extracted texture features using Maximum Response (MR) filters and input them into deep coupled autoencoders for classification. Other studies that utilized transfer learning methodologies can be found in \citep{Luz2018, Silva2018, Kumari2020a}. \cite{Proenca2018} utilized deep CNN to emphasize the importance of the periocular region for recognition by training the network with augmented periocular images having inconsistent iris and sclera regions. The training procedure causes the network to implicitly disregard the iris and sclera region. The authors in \citep{Zhao2018, Wang2021} integrated attention model to the deep architecture in order to highlight the significant regions (eyebrow and eye) of the periocular image. Some researchers utilized existing off-the-shelf CNN models to extract deep features at various convolutional levels \citep{HernandezDiaz2018,Kim2018,Hwang2020,Kumari2020a, Kumari2021a}. The authors in \citep{Zhang2018, Reddy2018b} proposed compact and custom deep learning models for use in mobile devices. Other deep learning-based methods include PatchCNN \citep{Reddy2018a}, “In-Set” CNN Iterative Analysis \citep{Proenca2019}, unsupervised convolutional autoencoders \citep{Reddy2019}, compact Convolutional Neural Network (CNN) \citep{Reddy2020}, VisobNet \citep{Ahuja2017}, semantics assisted CNN \citep{Zhao2017}, heterogeneity aware deep embedding \citep{Garg2018}, and Generalized Label Smoothing Regularization (GLSR)-trained networks \citep{Jung2020}. Deep learning approaches provide state-of-the-art recognition performance, but their performance are heavily data-driven.    

After the feature extraction step, some researchers further processed the feature vector, which generally includes the application of feature selection, subspace projection, or dimensional reduction \citep{Oh2012a, Joshi2014} techniques. These techniques aim to transform the feature set into a condensed representative feature set such that it improves the accuracy and reduces the computational complexity. Various post-processing techniques used in periocular recognition are Principal Component Analysis (PCA) \citep{Oh2012a}, Direct Linear Discriminant Analysis (DLDA) \citep{Joshi2014}, and Particle Swarm Optimization \citep{Silva2018}. Finally, the processed features are compared using similarity or dissimilarity metrics such as Bhattacharya distance \citep{Woodard2010a}, Hamming distance \citep{Oh2014}, I-Divergence metric \citep{Cao2014a}, Euclidean distance\citep{Ambika2016}, or Mahalanobis distance \citep{Nie2014}.

\begin{table*}[]
\caption{A list of feature extraction techniques under global, local, and deep learning categories, along with some representative research papers describing these techniques.}
\label{table:periocular-features}
\resizebox{\textwidth}{!}{%
\begin{tabular}{|l|l|l|l|}
\hline
\textbf{Features} & \textbf{References} & \textbf{Features} & \textbf{References} \\ \hline
\multicolumn{4}{|c|}{\textbf{Global Features}} \\ \hline
Local Binary Patterns (LBP) & \begin{tabular}[c]{@{}l@{}}\citep{Park2009,Bharadwaj2010,Xu2010,Oh2012a}\\ \citep{Padole2012, Santos2012,Uzair2013}\\ \citep{Mahalingam2014,Nie2014,Sharma2014,Bakshi2015} \\ \citep{JuefeiXu2010,Miller2010a,Santos2015} \\ \citep{JuefeiXu2012,Cao2014a}\end{tabular} & Bayesian Graphical Models (BGM) & \citep{Boddeti2011} \\ \hline
Gabor filters & \begin{tabular}[c]{@{}l@{}} \citep{JuefeiXu2010,AlonsoFernandez2012} \\ \citep{Cao2014a,Joshi2014}\\\citep{AlonsoFernandez2015a} \end{tabular} & Force Field Transform (FFT) & \citep{JuefeiXu2010} \\ \hline
Probabilistic Deformation Models (PDM) & \citep{Ross2012,Smereka2013} & GIST perceptual descriptors & \citep{Bharadwaj2010} \\ \hline
Binarized Statistical Image Features (BSIF) & \citep{Raghavendra2013,Raja2014a} & Leung-Mallik filters (LMF) & \citep{Tan2012b} \\ \hline
Joint Dictionary-based Sparse Representation & \citep{Raghavendra2013,Jillela2014,Moreno2016} & Laplacian of Gaussian (LoG) & \citep{JuefeiXu2010} \\ \hline
Discrete Wavelet Transform (DWT) & \citep{JuefeiXu2010,Joshi2014} & Laws masks & \citep{JuefeiXu2010} \\ \hline
Histogram of Oriented Gradients (HOG) & \citep{Park2009, Park2011,Algashaam2017b} & Discrete Cosine Transform (DCT) & \citep{JuefeiXu2010} \\ \hline
Phase Intensive Global Pattern (PIGP) & \citep{Smereka2013,Bakshi2015} & Normalized Gradient Correlation (NGC) & \citep{Jillela2014} \\ \hline
Structured Random Projections (SRP) & \citep{Oh2014} & Walsh masks & \citep{JuefeiXu2010} \\ \hline
Shape-based features & \begin{tabular}[c]{@{}l@{}}\citep{Dong2011,Proenca2014a} \\ \citep{,Le2014, Ambika2016} \end{tabular} & Gaussian Markov random field & \citep{Smereka2015} \\ \hline
Maximum Response (MR) & \citep{Raghavendra2016a} & 2D and 3D features & \citep{Chen2015} \\ \hline
Color-based features & \citep{Woodard2010a, Lyle2012, Moreno2016} & Genetic and Evolutionary Feature Extraction & \citep{Adams2010a} \\ \hline
\multicolumn{4}{|c|}{\textbf{Local Features}} \\ \hline
Scale Invariant Feature Transformation (SIFT) & \begin{tabular}[c]{@{}l@{}}\citep{Xu2010,Park2011,Padole2012}\\ \citep{Ross2012,Santos2012,Smereka2013} \\ \citep{AlonsoFernandez2014b,Ahuja2016a} \end{tabular} & \begin{tabular}[c]{@{}l@{}}Binary Robust Invariant Scalable Key points \\ (BRISK)\end{tabular} & \citep{Mikaelyan2014} \\ \hline
Speeded-up Robust Features (SURF) & \begin{tabular}[c]{@{}l@{}}\citep{JuefeiXu2010,Xu2010}\\\citep{,Bakshi2015,Raja2015a} \end{tabular} & Phase Intensive Local Pattern (PILP) & \citep{Bakshi2015} \\ \hline
\begin{tabular}[c]{@{}l@{}}Symmetry Assessment by Feature Expansion \\  (SAFE) \end{tabular}& \citep{Mikaelyan2014,AlonsoFernandez2015a} & Oriented FAST and Rotated BRIEF (ORB) & \citep{Mikaelyan2014} \\ \hline
\multicolumn{4}{|c|}{\textbf{Deep Learning Techniques}} \\ \hline
Off-the-shelf CNN Features & \begin{tabular}[c]{@{}l@{}}\citep{HernandezDiaz2018,Kim2018} \\ \citep{Hwang2020, Kumari2021a} \end{tabular} & \begin{tabular}[c]{@{}l@{}}Convolutional Restricted Boltzman \\Machines (CRBM) \end{tabular}& \citep{Nie2014} \\ \hline
Transfer Learning & \begin{tabular}[c]{@{}l@{}}\citep{Ahuja2017,Luz2018}\\\citep{Silva2018, Kumari2020a}\end{tabular} & PatchCNN & \citep{Reddy2018a} \\ \hline
Deep Sparse Filters & \citep{Raja2016a, Raja2020} & “In-Set” CNN Iterative Analysis & \citep{Proenca2019} \\ \hline
Custom and Compact CNNs & \citep{Reddy2018b,Zhang2018,Reddy2020} & Semantics Assisted CNN & \citep{Zhao2017} \\ \hline
Autoencoders & \citep{Raghavendra2016a,Reddy2019} & Heterogeneity Aware Deep Embedding & \citep{Garg2018} \\ \hline
Attention Models & \citep{Zhao2018, Wang2021} & \begin{tabular}[c]{@{}l@{}} Generalized Label Smoothing \\Regularization (GLSR)\end{tabular} & \citep{Jung2020} \\ \hline
\end{tabular}
}
\end{table*}

\section{Periocular Recognition in Different Spectra}
\label{sec:Periocular-Spectrums}
Different imaging spectra have been described in the literature for capturing the periocular region, including Near-Infrared (NIR), Visible (VIS), Short Wave Infrared (SWIR), Middle Wave Infrared (MWIR), and Long Wave Infrared (LWIR). The most commonly used imaging spectra are NIR and VIS. This is because most research in periocular biometrics is based on face images (VIS) or iris images (NIR). Further, even as a standalone biometric, periocular images are captured using existing face or iris sensors. The NIR spectrum, which operates in the 700-900nm range, predominantly captures the iris pattern, eye shape, outer and inner corner of the eye, eyelashes, eyebrows, and eyelids. Often there is saturation in the area around the eye, skin, and cheek regions. On the other hand, the VIS spectrum (400-700nm) captures textural details of the periocular skin region, conjunctiva vasculature, eye shape, eyelashes, eyebrows, and eyelids. The VIS imaging fails to capture the textural nuances of the iris pattern, especially for dark-colored irides. Examples of periocular recognition techniques in the NIR spectrum are \citep{Monwar2013,Uzair2013,Hwang2020, Mikaelyan2014}, and in the VIS spectrum are \citep{Adams2010a,Bharadwaj2010,Park2009,JuefeiXu2010, Miller2010b, Woodard2010b, Xu2010, Park2011, Oh2012a,Padole2012, Santos2012, Joshi2014, Nie2014,Proenca2014c, Proenca2014b, Bakshi2015, Santos2015, HernandezDiaz2018, Luz2018, Reddy2019}. \cite{Rattani2017a} provided a detailed survey of ocular techniques in the VIS spectrum. The researchers in \citep{Hollingsworth2010, Smereka2017} suggested that VIS images provide more discriminative information for periocular recognition compared to NIR images. \cite{Hollingsworth2012} made the same conclusion using human volunteers. The authors in \citep{AlonsoFernandez2012,Ross2012,AlonsoFernandez2015a, Smereka2015, Ambika2016, Zhao2017} proposed periocular recognition techniques that can be applied to both NIR and VIS images. Other researchers \citep{Algashaam2017a, Vetrekar2018, Ipe2020} fused information obtained from both NIR and VIS images. Table \ref{table:periocular-spectrum} provides a summary (features extraction, datasets, and performance) of various techniques applied on NIR, VIS, both spectrum, and multi-spectral (fusion of NIR and VIS) images. 

A vast amount of research has also focused on cross-spectrum matching, where enrolled images are in one spectrum, while probe images are in another spectrum. The cross-spectrum evaluation scenario implicitly encapsulates the cross-sensor scenario (enrolled and probes images are from different sensors) as well. Examples of papers discussing the cross-spectrum scenario are \citep{Cao2014a, Sharma2014, Ramaiah2016, Behera2017,Raja2017, HernandezDiaz2019, AlonsoFernandez2020, Behera2020,HernandezDiaz2020,Zanlorensi2020b,Vyas2022}. \cite{Behera2019} provided a detailed survey on cross-spectrum periocular recognition. A more difficult evaluation scenario is when testing is performed on different datasets (cross-dataset) as it has to account for the variations due to different sensors, data acquisition environments, and subject population. Examples of cross-dataset evaluation can be found in \citep{Reddy2019, Reddy2020}. Table \ref{table:periocular-cross} summarizes various cross-spectrum and cross-dataset techniques. The cross-sensor techniques are mainly evaluated on different mobile devices, so we provide these details in Section \ref{sec:Periocular-MobileDevices} (Periocular Recognition on Mobile Devices).

\begin{table*}[]
\caption{A chronological overview (description, datasets, and performance) of periocular techniques utilizing NIR, VIS, or multispectral images. Here, RR is Recognition Rate, EER is Equal Error Rate, TMR is True Match Rate, FRR is False Rejection Rate, and FAR is False Acceptance Rate. The acronyms used in the `Description' column are defined in the text or in the referenced papers.}
\label{table:periocular-spectrum}
\resizebox{\textwidth}{!}{%
\begin{tabular}{|l|l|l|}
\hline
\textbf{Paper} & \textbf{Description} & \textbf{Datasets and Performance} \\ \hline
\multicolumn{3}{|c|}{\textbf{NIR Spectrum}} \\ \hline
\citep{Uzair2013} & \begin{tabular}[c]{@{}l@{}}Formulate as an image set classification problem, \\ where each image set corresponds to single subject\end{tabular} & MBGC: RR is 97.70\% \\ \hline
\citep{Monwar2013} & \begin{tabular}[c]{@{}l@{}}PDM, modified SIFT, GOH features\\ Fusion: Highest rank, borda count, plurality voting, \\ markov chain rule at rank-level\end{tabular} & FOCS: RR is 99.2\% \\ \hline
\citep{Mikaelyan2014} & Symmetry patterns & BioSec: EER is 10.75\% \\ \hline
\citep{Hwang2020} & \begin{tabular}[c]{@{}l@{}}Mid-level CNN features (plain CNN, ResNet, \\ deep plane CNN, and deep ResNet) + Features selection\end{tabular} & \begin{tabular}[c]{@{}l@{}}Proprietary: EER is 11.51\% \\ CASIA-Iris-Lamp: EER is 0.64\%\end{tabular} \\ \hline
\multicolumn{3}{|c|}{\textbf{Visible Spectrum}} \\ \hline
\citep{Park2009} & HOG, LBP, SIFT & FRGC: RR is 79.49\%(SIFT) \\ \hline
\citep{Xu2010} & Comparison of different features and their fusion & FRGC: TMR of 61.2\% 0.1\% FMR \\ \hline
\citep{Adams2010a} & GEFE+LBP & \begin{tabular}[c]{@{}l@{}}FRGC: RR is 92.16\% \\ FERET: RR is 85.06\%\end{tabular} \\ \hline
\citep{JuefeiXu2010} & \begin{tabular}[c]{@{}l@{}}LBP, WLBP, SIFT, DCT, Gabor filters, Walsh masks, \\ DWT, SURF Law Masks, Force Fields, LoG\end{tabular} & \begin{tabular}[c]{@{}l@{}}FRGC: RR is 53.2\%(LBP+DWT) \\ FG-NET: RR is 53.1\%(LBP+DCT)\end{tabular} \\ \hline
\citep{Park2011} & Fusion of HOF, LBP and SIFT & FRGC: RR is 87.32\% \\ \hline
\citep{Padole2012} & HOG, LBP, SIFT & UBIPr: EER is $\sim$20\%(HOG + LBP + SIFT) \\ \hline
\citep{Santos2012} & LBP, SIFT & UBIRIS v2: EER is 31.87\% and RR is 56.4\% \\ \hline
\citep{Joshi2014} & Gabor-PPNN, DWT, LBP, HOG & \begin{tabular}[c]{@{}l@{}}MBGC: EER is 6.4\%, GTDB: EER is 5.9\%, \\ IITK: EER is 15.5\%, PUT: EER is 4.8\%\end{tabular} \\ \hline
\citep{Nie2014} & PCA to: CRBM, SIFT, LBP, HOG & UBIPr: EER is 6.4\% and RR is 50.1\% \\ \hline
\citep{Proenca2014c} & GC-EGM to: LBP + HOG + SIFT & FaceExpressUBI: EER is 16\% \\ \hline
\citep{HernandezDiaz2018} & \begin{tabular}[c]{@{}l@{}}Fusion of off-the-shelf CNN (AlexNet, GoogLeNet,\\  ResNet, and VGG) features and traditional features\end{tabular} & \begin{tabular}[c]{@{}l@{}}UBIPr: EER of 5.1\% and\\  FRR is 11.3\% at 1\% FAR\end{tabular} \\ \hline
\citep{Jung2020} & Generalized label smoothing regularization-trained networks & \begin{tabular}[c]{@{}l@{}}ETHNIC, PUBFIG, FACESCRUB, AND IMDB \\ WIKI: avg.RR is 88.7\% and EER of 10.4\%\end{tabular} \\ \hline
\multicolumn{3}{|c|}{\textbf{NIR and VIS Spectrum}} \\ \hline
\citep{Woodard2010b} & Tessellated image + Histograms of texture and color & FRGC (VIS): RR is 91\%, MBGC (NIR): RR is 87\% \\ \hline
\citep{Ross2012} & Fusion of GOH, PDM, SIFT features at the score-level & \begin{tabular}[c]{@{}l@{}}FOCS (NIR): EER is 18.8\%, FRGC (VIS): EER is 1.59\%\end{tabular} \\ \hline
\citep{AlonsoFernandez2015a} & Gabor features & \begin{tabular}[c]{@{}l@{}}4 NIR datasets: Accuracy is 97\% \\ 2 VIS datasets: Accuracy is 27\%\end{tabular} \\ \hline
\citep{Bakshi2015} & Raw pixels, LBP, PCA, LBP + PCA & \begin{tabular}[c]{@{}l@{}}MGBC: NIR- RR is 99.8\%, VIS- RR is 98.5\% \\ CMU Hyperspectral: RR is 97.2\%, UBIPr: RR is 99.5\%\end{tabular} \\ \hline
\citep{Ambika2016} & Laplace–Beltrami based shape features & \begin{tabular}[c]{@{}l@{}}CASIA FV1: accuracy is 95\%, Basel 3D: Accuracy is 97\%  \\ 3D periocular: Accuracy is 97.5\%\end{tabular} \\ \hline
\citep{Smereka2015} & Periocular probabilistic deformation models & 2 NIR and 3 VIS images datasets \\ \hline
\citep{Zhao2017} & Semantics-assisted convolutional neural networks & \begin{tabular}[c]{@{}l@{}}UBIRIS.V2: RR is 82.43\%, FRGC: RR is 91.13\%, \\ FOCS: RR is 96.93\%, CASIA.v4-distance: RR is 98.90\%\end{tabular} \\ \hline
\multicolumn{3}{|c|}{\textbf{Multi-spectrum}} \\ \hline
\citep{Algashaam2017a} & Multimodal compact multi-linear pooling feature fusion & IMP: Accuracy is 91.8\% \\ \hline
\citep{Vetrekar2018} & HOG, GIST, Log-Gabor transform and BSIF + CRC & Proprietary: RR is 96.92\% \\ \hline
\citep{Ipe2020} & Fusion of the off-the-shelf CNN feature & IMP: Accuracy is 97.14\% \\ \hline
\end{tabular}
}
\end{table*}

\begin{table*}[t]
\caption{A chronological overview (description, datasets, and performance) of periocular techniques working under cross-spectrum and cross-dataset scenarios. Here, GAR is Genuine Acceptance Rate, GMR is Genuine Match Rate, FMR is False Match Rate, and $d^\prime$ is the separation between the mean of genuine and impostor distributions. The acronyms used in the `Description' column are defined in the text or in the referenced papers.}
\label{table:periocular-cross}
\resizebox{\textwidth}{!}{%
\begin{tabular}{|l|l|l|}
\hline
\textbf{Paper} & \textbf{Description} & \textbf{Datasets and Performance} \\ \hline
\multicolumn{3}{|c|}{\textbf{Cross-spectrum}} \\ \hline
\citep{Cao2014a} & \begin{tabular}[c]{@{}l@{}}Gabor LBP, Generalized LBP, Gabor Weber descriptors\end{tabular} & \begin{tabular}[c]{@{}l@{}}Pre-Tinders,TINDERS, PCSO (GAR at 0.1 FAR): SWIR-VIS: 0.75, \\ NIR-VIS: 0.35, MWIR-VIS: 0.35\end{tabular} \\ \hline
\citep{Sharma2014} & Combined neural network architecture & \begin{tabular}[c]{@{}l@{}}IMP (GAR @ 1\% FAR): VIS-NV: 71.93\%, VIS-NIR:47.08\%, NV-NIR:48.21\%\end{tabular} \\ \hline
\citep{Ramaiah2016} & Three patch LBP + MRF & \begin{tabular}[c]{@{}l@{}}(GAR @ 0.1\% FAR) IMP (NIR-VIS): 18.35\%, PolyU (NIR-VIS): 73.2\% \end{tabular} \\ \hline
\citep{Raja2017} & \begin{tabular}[c]{@{}l@{}}Steerable pyramids + SVM + Fusion of different scales\end{tabular} & CROSS-EYED2016 (NIR-VIS): GMR is 100\% at 0.01\% FMR \\ \hline
\citep{Behera2017} & \begin{tabular}[c]{@{}l@{}}Difference of Gaussian + HOG + Cosine similarity\end{tabular} & \begin{tabular}[c]{@{}l@{}}IMP (NIR-VIS):GAR is 25.03\% at 0.1 FAR, EER is 45.29\% \\ PolyU: GAR is 83.12\% at 0.1 FAR, EER is 13.87\% \\ CROSS-EYED2016: GAR is 89.27\% at 0.1 FAR and EER is 13.22\%\end{tabular} \\ \hline
\citep{HernandezDiaz2019} & \begin{tabular}[c]{@{}l@{}}ResNet101 features + Chi-square distance, Cosine similitude\end{tabular} & \begin{tabular}[c]{@{}l@{}}IMP (EER, GAR at 1\% FAR): VIS-NV: (5.13\%, 88.19)\\ VIS-NIR: (5.19\%, 88.13); NIR-NV: (10.19\%, 81.55)\end{tabular} \\ \hline
\citep{HernandezDiaz2020} & \begin{tabular}[c]{@{}l@{}}Convert NIR and VIS images using cGAN +  CNN features\end{tabular} & PolyU (NIR-VIS): EER is 1\%, GAR is 99.1\% at FAR=1\% \\ \hline
\citep{Zanlorensi2020b} & Fine-tune CNN models ( VGG16, ResNet-50) & PolyU (NIR-VIS): EER is 0.35\% and d’ is 7.75 \\ \hline
\citep{Behera2020} & Variance-guided attention-based twin deep network & \begin{tabular}[c]{@{}l@{}}PolyU: EER is 6.38\%, GAR is 96.17\% at 10\% FAR \\ CROSS-EYED2016: EER is 2.36\%, and GAR is 99.70\% at 10\% FAR \\ IMP (EER, GAR at 10\%FAR): VIS-NV: (9.71\%, 90.62\%) \\ VIS-NIR: (13.59\%, 82.49\%), NIR-NV: (7.06\%, 95.17\%)\end{tabular} \\ \hline
\citep{AlonsoFernandez2020} & \begin{tabular}[c]{@{}l@{}}Fuse HOG, LBP, SIFT, Symmetry Descriptors, Gabor, \\ Steerable Pyramidal Phase, VGG-Face, Resnet101, \\ and Densenet201 features using LLR at score-level\end{tabular} & \begin{tabular}[c]{@{}l@{}}CROSS-EYED2016: EER is 0.2\%, FRR is 0.47\% at 0.01\% FAR \\ VSSIRIS: EER is 0.2\%, FRR 0.3\% at 0.01\% FAR\end{tabular} \\ \hline
\multicolumn{3}{|c|}{\textbf{Cross-datasets}} \\ \hline
\citep{Reddy2019} & Unsupervised convolutional autoencoders & \begin{tabular}[c]{@{}l@{}}Train: UBIRIS-V2, UBIPr, MICHE; Test: VISOB: EER is 12.23\%\end{tabular} \\ \hline
\citep{Reddy2020} & CNN features & \begin{tabular}[c]{@{}l@{}}Train: VISOB, Test:UBIRIS-V2: EER is 7.65\%,\\ UBIPr: EER is 3.87\% CROSS-EYED2016: EER is 0.94\%,\\ CASIA-TWINS: EER is 9.41\% FERET: EER is 0.06\%\end{tabular} \\ \hline
\end{tabular}
}
\end{table*}

\section{Periocular Fusion with Other Modalities}
\label{sec:Periocular-Fusion}
Simultaneous acquisition of periocular with the iris modality, and its complementary nature with respect to iris, has motivated researchers to fuse periocular with iris to improve the overall recognition performance. The authors in \citep{Woodard2010a, Ross2012} proposed the fusion of periocular with iris to improve the performance when acquired iris images are of low quality due to partial occlusions, specular reflections, off-axis gaze, motion and spatial blur, non-linear deformations, contrast variations, and illumination artifacts. The fusion is also helpful when iris images are captured from a distance as the periocular region is relatively stable even at a distance \citep{Tan2012b}. It is also advantageous when iris images are acquired in the visible spectrum \citep{Santos2012, Tan2013, Proenca2014a, Jain2015, Silva2018}, or using mobile devices \citep{Santos2015, Ahuja2016b}. The iris texture is better discernible in NIR illumination, whereas periocular features become more perceptible in VIS illumination \citep{AlonsoFernandez2015a, AlonsoFernandez2015b}. The overall performance obtained on the fusion of iris and periocular traits is generally better than using the iris only as shown in \citep{Komogortsev2012,Raghavendra2013, Raja2014a, Ahmed2017, Verma2016}. The fusion of iris and periocular is mainly performed at the score-level \citep{Woodard2010a, Tan2012a, Tan2012b, Raghavendra2013,Tan2013, Proenca2014a, AlonsoFernandez2015b,Jain2015, Santos2015, Verma2016, Ahuja2016b, Algashaam2021}, though there is some work on feature-level \citep{Jain2015, Stokkenes2017, Silva2018} and decision-level \citep{Santos2012} fusion also. \cite{Ogawa2021} proposed Multi Modal Selector that adaptively selects a iris and periocular classifier useful for human recognition. 

The fusion of periocular with the face modality is also a viable option as periocular is a part of the face, and no additional acquisition is required. Though the periocular region is already accounted in face recognition as a part of the face, isolating the periocular and performing region-speciﬁc feature extraction provides an overall improvement in recognition performance. The fusion of face with periocular is also beneficial when face images are occluded, having large pose variations, or captured at a very close distance (e.g., a selfie). The work of periocular fusion with  face in the context of plastic surgery \citep{Jillela2012}, gender transformation \citep{Mahalingam2014} and mobile devices \citep{Raja2015b, Pereira2015} shows improved recognition accuracy. Table \ref{table:periocular-fusion} summarizes various techniques that fuse periocular with iris and face modalities. In another research work, \cite{Oh2014} fused periocular features (structured random projections) with binary sclera features at the score-level for identity verification. \cite{Nigam2015} provided a detailed survey on the fusion of various ocular biometrics. 

\begin{table*}[]
\caption{A chronological overview (description, datasets, and performance) of periocular techniques focusing on the fusion of periocular with iris and face modalities. Here, AUC is Area Under the Curve. The acronyms used in the `Description' column are defined in the text or in the referenced papers.}
\label{table:periocular-fusion}
\resizebox{\textwidth}{!}{%
\begin{tabular}{|l|l|l|}
\hline
\textbf{Paper} & \textbf{Description} & \textbf{Datasets and Performance} \\ \hline
\multicolumn{3}{|c|}{\textbf{Fusion of Iris and Periocular}} \\ \hline
\citep{Woodard2010a} & \begin{tabular}[c]{@{}l@{}}Iris: Gabor features,  Periocular: LBP \\ Fusion: Weighted sum at score-level\end{tabular} & \begin{tabular}[c]{@{}l@{}}MBGC: EER is 0.18, RR is 96.5\%\end{tabular} \\ \hline
\citep{Santos2012} & \begin{tabular}[c]{@{}l@{}}Iris: Wavelets, Periocular: LBP, SIFT \\ Fusion: Logistic regression at decision-level\end{tabular} & \begin{tabular}[c]{@{}l@{}}NICE.II: EER is 18.48, AUC is 0.90, RR is 74.3\%\end{tabular} \\ \hline
\citep{Tan2012a} & \begin{tabular}[c]{@{}l@{}}Iris: Ordinal measures and color analysis \\ Periocular: Texton representation and semantic information \\ Fusion: Weighted sum rule at score-level\end{tabular} & \begin{tabular}[c]{@{}l@{}}USIRISv2: d’ is 2.57, EER is 12\%\end{tabular} \\ \hline
\citep{Tan2012b} & \begin{tabular}[c]{@{}l@{}}Iris: Log-Gabor features \\ Periocular: SIFT, Leung-Malik filters, LBP \\ Fusion: Weighted sum rule at score-level\end{tabular} & CASIA-IrisV4-distance: RR is 84.5\% \\ \hline
\citep{Tan2013} & \begin{tabular}[c]{@{}l@{}}Iris: Log-Gabor features \\ Periocular: DSIFT, GIST, LBP, HOG, LMF \\ Fusion: Weighted sum rule at score-level\end{tabular} & \begin{tabular}[c]{@{}l@{}}UBIRIS V2: RR is 39.6\% \\ FRGC: RR is 59.9\% \\ CASIA-IrisV4-Distance: RR is 93.9\%\end{tabular} \\ \hline
\citep{Raghavendra2013} & \begin{tabular}[c]{@{}l@{}}Iris, Periocular: LBP-SRC, Fusion: Weighted sum at score-level\end{tabular} & Proprietary: EER is 0.81\% \\ \hline
\citep{Proenca2014a} & \begin{tabular}[c]{@{}l@{}}Iris: Multi-lobe differential filters \\ Eyelids, Eyelashes, Skin: shape and LBP features \\ Fusion: Product, sum, min and max rule at score-level\end{tabular} & \begin{tabular}[c]{@{}l@{}}UBIRISv2: d’ is 2.97 and AUC is 0.96 \\ FRGC: d’ is 3.02 and AUC is 0.97\end{tabular} \\ \hline
\citep{Santos2015} & \begin{tabular}[c]{@{}l@{}}Iris: Gabor features, Periocular: SIFT, GIST, LBP, HOG, ULBP \\ Fusion: ANN at score-level\end{tabular} & \begin{tabular}[c]{@{}l@{}}CSIP: d' is 2.501, AUC is 0.943,\\ EER is 0.131\end{tabular} \\ \hline
\citep{Jain2015} & \begin{tabular}[c]{@{}l@{}}Iris, Periocular: LBP, SIFT, GIST \\ Fusion: Feature-level (context-switching), score-level (sum)\end{tabular} & \begin{tabular}[c]{@{}l@{}}UBIRISv2: RR-10 is 76.16 \%\\ FRGC: RR-10 is 75.4 \%\end{tabular} \\ \hline
\citep{AlonsoFernandez2015b} & \begin{tabular}[c]{@{}l@{}}Iris: Log-Gabor filters, DCT, SIFT \\ Periocular: Symmetry patterns, gabor features, SIFT \\ Fusion: Logistic regression at score-level\end{tabular} & \begin{tabular}[c]{@{}l@{}}(EER) BioSec: 0.75\%, MobBIO: 6.75\% \\ CASIA-Iris Inverval v3: 0.51\% \\ IIT Delhi v1.0: 0.38\%, UBIRIS v2: 15.17\%\end{tabular} \\ \hline
\citep{Verma2016} & \begin{tabular}[c]{@{}l@{}}Iris: Gabor features,  Periocular: PHOG, GIST \\ Fusion: Random decision forest at score-level\end{tabular} & \begin{tabular}[c]{@{}l@{}}CASIA-IrisV4-distance: GMR is 61\% at 0.1\% FMR\\  FOCS: GMR is 21\% at 0.1\% FMR\end{tabular} \\ \hline
\citep{Ahuja2016b} & \begin{tabular}[c]{@{}l@{}}Iris: RootSIFT, Periocular: Deep features \\ Fusion: Mean rule and linear regression at score-level\end{tabular} & MICHE-II: AUC is 0.985 and EER is 0.057 \\ \hline
\citep{Ahmed2017} & \begin{tabular}[c]{@{}l@{}}Iris: Gabor features, Periocular: Multi-Block Transitional LBP \\ Fusion: Weighted sum rule at score-level\end{tabular} & \begin{tabular}[c]{@{}l@{}}MICHE II: EER is 1.22\%, \\ FRR is 2.56\% at FAR\\ RR is 100\%\end{tabular} \\ \hline
\citep{Zhang2018} & \begin{tabular}[c]{@{}l@{}}Iris, Periocuar: CNNs with max out units\\ Fusion: Weighted concatenation at the feature-level\end{tabular} & \begin{tabular}[c]{@{}l@{}}CASIA-Iris-MobileV1.0: EER is 0.60\%, \\ FNMR is 2.32\% at 0.001\% FMR\end{tabular} \\ \hline
\citep{Silva2018} & \begin{tabular}[c]{@{}l@{}}Iris, Periocular: Deep features\\ Fusion: Particle swarm optimization at feature-level\end{tabular} & UBIRISv2: d' is 3.45 and EER is 5.55\% \\ \hline
\multicolumn{3}{|c|}{\textbf{Fusion of Periocular and Face}} \\ \hline
\citep{Jillela2012} & \begin{tabular}[c]{@{}l@{}}Face: Verilook and PittPatt Scores, Ocular: SIFT, LBP \\ Fusion: Mean rule at score-level fusion\end{tabular} & Plastic surgery database: RR is 87.4\% \\ \hline
\citep{Pereira2015} & \begin{tabular}[c]{@{}l@{}}Periocular: Tessellated images + DCT + GMM \\ Face: Inter-session variability modeling + GMM \\ Fusion: Linear logistic regression at score-level fusion\end{tabular} & \begin{tabular}[c]{@{}l@{}}MOBIO: HTER is 6.58\% \\ CPqD Biometric: HTER is 3.87\%\end{tabular} \\ \hline
\citep{Raja2015a} & \begin{tabular}[c]{@{}l@{}}Iris: Gabor features,  Periocular, Face: SIFT, SURF, BSIF\\ Fusion: Min, max, product, weighted sum at score-level\end{tabular} & Proprietary: EER of 0.68\% \\ \hline
\citep{Stokkenes2017} & \begin{tabular}[c]{@{}l@{}}Face, Periocular: BSIF + Bloom filters  \\ Fusion: XOR operation, concatenation at feature-level\end{tabular} & \begin{tabular}[c]{@{}l@{}}Proprietary: GMR is 88.54\% at 0.01\% FMR \\ EER is 2.05\%\end{tabular} \\ \hline
\end{tabular}
}
\end{table*}

\section{Periocular Recognition on Mobile Devices}
\label{sec:Periocular-MobileDevices}
The extensive usage of mobile devices motivates the need for human authentication on mobile devices for various purposes, such as access control, digital payments, or mobile banking. Several mobile devices are now emerging with integrated biometric sensors -- iPhone 12 has a Touch ID fingerprint sensor and Face ID cameras, and the Samsung Galaxy S20 series has an in-display fingerprint sensor and an iris scanner. Periocular images are generally acquired using the front or rear camera of mobile devices in the visible spectrum. The challenges in mobile biometrics are low-quality input images and relatively limited computational power. The low-quality images are due to hardware limitations and less constrained capturing environments. \cite{Raja2014b} explored periocular recognition on smart devices using well known feature extraction techniques (SIFT, SURF, and BSIF) and achieved a Genuine Match Rate (GMR) of 89.38\% at 0.01\% False Match Rate (FMR). There is some work on NIR images captured from mobile devices \citep{Bakshi2018, Zhang2018}. \cite{Bakshi2018} utilized a reduced version of Phase Intensive Local Pattern (PILP) features, whereas \citep{Zhang2018} fused compact CNN features of iris and periocular through a weighted concatenation. Majority of the periocular-based mobile biometrics are performed on VIS images \citep{Pereira2015, Raja2015a, Ahuja2016a,Keshari2016,Raja2016a, Raghavendra2016a, Ahmed2017, Ahuja2017,Rattani2017b,Stokkenes2017, Boutros2020b, Krishnan2020,Raja2020}. \cite{Keshari2016} investigated periocular recognition on pre- and post-cataract surgery mobile images. \cite{Krishnan2020} investigated the fairness of mobile ocular biometrics methods across gender. The work in \citep{Pereira2015,Raja2015a,Ahmed2017, Zhang2018} used fusion of different modalities for mobile biometrics -- \cite{Raja2015a} fused iris, face and periocular modalities, \citep{Pereira2015} combined face and periocular, whereas \citep{Santos2015,Ahmed2017,Zhang2018} combined iris and periocular. Recent work on mobile biometrics used deep learning features \citep{Raja2016a, Raghavendra2016a, Ahuja2017, Rattani2017b, Raja2020}. \cite{Boutros2020b} verified an individual wearing Head Mounted Display (HMD) using four handcrafted feature extraction methods and two deep-learning strategies. Generalizability across different mobile sensors (cross-sensor) are also evaluated in \citep{Santos2015, Raja2016c, Raja2016d, Garg2018, Reddy2018a, AlonsoFernandez2020}. Table \ref{table:periocular-mobile} provides a brief description of various mobile-based periocular recognition techniques.

\begin{table*}[t]
\caption{A chronological overview (description, datasets, and performance) of periocular techniques utilizing images acquired using the sensors and cameras in a mobile device such as a smartphone. Here, HTER is Half Total Error Rate. The acronyms used in the `Description' column are defined in the text or in the referenced papers.}
\label{table:periocular-mobile}
\resizebox{\textwidth}{!}{%
\begin{tabular}{|l|l|l|}
\hline
\textbf{Paper} & \textbf{Description} & \textbf{Datasets and Performance} \\ \hline
\citep{JuefeiXu2012} & \begin{tabular}[c]{@{}l@{}}Walsh-Hadamard transform encoded LBP\\ + Kernel class-dependence feature analysis\end{tabular} & Compass: GAR is 60.7\% at 0.1\% FAR \\ \hline
\citep{Raja2014b} & \begin{tabular}[c]{@{}l@{}}SIFT, SURF and BSIF\\ + Nearest Neighbors (SIFT and SURF), \\ Bhattacharyya distance(BSIF)\end{tabular} & Proprietary: GAR is 89.38\% at 0.01\% FAR \\ \hline
\citep{Ahuja2016a} & \begin{tabular}[c]{@{}l@{}}SURF + Multinomial Naive Bayes learning \\ + Pyramid-up topology using Dense SIFT + RANSAC\end{tabular} & VISOB: RR is 48.76\%-79.49\% \\ \hline
\citep{Keshari2016} & \begin{tabular}[c]{@{}l@{}}Dense SIFT, Gabor, Scattering Network features + \\ PCA + LDA + Cosine similarity weighted sum\end{tabular} & \begin{tabular}[c]{@{}l@{}}IMP: RR-10 is 69\%\\ GAR is 24\% at 1\% FAR\end{tabular} \\ \hline
\citep{Raghavendra2016a} & \begin{tabular}[c]{@{}l@{}}Maximum Response filters + Deeply coupled autoencoders\end{tabular} & VISOB: GMR of 93.98\% at 0.001 FMR \\ \hline
\citep{Raja2016d} & \begin{tabular}[c]{@{}l@{}}Laplacian decomposition + GLCM +\\ STFT + Histogram features of freq. response + SRC\end{tabular} & \begin{tabular}[c]{@{}l@{}}MICHE I: Cross-camera EER is 7.53\%\\  Cross-sensor EER is 6.38\%\end{tabular} \\ \hline
\citep{Ahuja2017} & Hybrid CNN model + Mean fusion at the score-level & \begin{tabular}[c]{@{}l@{}}VISOB: GMR is 99.5\% at 0.001\% FMR \\ MICHE-II: AUC of 98.6\%\end{tabular} \\ \hline
\citep{Rattani2017b} & Fine-tuned VGG-16, VGG-19, InceptionNet, ResNet & VISOB: TMR is 100\% at 0.001\% FMR \\ \hline
\citep{Garg2018} & Heterogeneity aware loss function in deep network & \begin{tabular}[c]{@{}l@{}}(RR) CSIP (cross-sensor): 89.53\%,\\ IMP: 61.20\%, VISOB (cross-spectrum): 99.41\%\end{tabular} \\ \hline
\citep{Reddy2018a} & Patch-based OcularNet & \begin{tabular}[c]{@{}l@{}} (EER) VISOB: 1.17\%, UBIRIS-I: 9.86\%,\\ UBIRIS-II: 9.77\%, CROSS-EYED2016:  14.95\%\end{tabular} \\ \hline
\citep{Raja2020} & \begin{tabular}[c]{@{}l@{}}Deep Sparse Features, Deep Sparse Time \\ Frequency Features + CRC classification\end{tabular} & \begin{tabular}[c]{@{}l@{}}(GMR at 0.01\% FMR) VISPI: 99.80\%, \\ MICHE-I: 100\%, VISOB: 98.78\%\end{tabular} \\ \hline
\citep{Boutros2020b} & \begin{tabular}[c]{@{}l@{}} Periocular, Iris:  Hand-crafted and deep features \\ + Synthesize identity-preserved periocular images\end{tabular} & \begin{tabular}[c]{@{}l@{}}OpenEDS: EER (iris) is 6.35\%  \\ EER (periocular) is 5.86\%\end{tabular} \\ \hline
\end{tabular}
}
\end{table*}

\section{Specific Applications}
\label{sec:Periocular-OtherApplications}

\begin{enumerate}
\item \textbf{Soft-biometrics from Periocular Region}: Soft-biometrics refer to attributes used to classify individuals in broad categories such as gender, ethnicity, race, age, height, weight, or hair color. The periocular region has also been used for automatically estimating age, gender, ethnicity, and facial expression information. An exploration of gender information contained in the periocular region is performed in \citep{Merkow2010, Lyle2012, Bobeldyk2016, CastrillonSantana2016, Tapia2019b}. \cite{Tapia2019b} synthesized NIR periocular images using a conditional GAN based on gender information, and then identify gender using the synthesized periocular images. The work in \citep{ Lyle2012, Woodard2017} extracted race information from the periocular region, while the work in \citep{Rattani2017d} determined an individual's age from the periocular region. \cite{AlonsoFernandez2018} investigated the feasibility of using the periocular region for facial expression recognition.  

\item \textbf{Long Distance Recognition}: \cite{Bharadwaj2010} showed the degradation of iris recognition performance with an increase in standoff distance and suggested the use of the periocular region on long-distance images. The authors in \citep{Tan2012b, Verma2016} proposed fusion approaches (iris and periocular) for human recognition at a distance (NIR images). \cite{Kim2018} presented CNN-based periocular recognition in a surveillance environment.

\item \textbf{Face Generation from Periocular Region}: \cite{JuefeiXu2014b, JuefeiXu2016} recreated the entire face from the periocular region alone using dictionary learning algorithms, while \cite{UdDin2020} proposed a GAN-based method to regenerate the masked part of the face. \cite{Li2020} utilized de-occlusion distillation framework to recover face content under the mask.

\item \textbf{Cross-modal Recognition (Face and Iris)}: \cite{Jillela2014} presented the challenging problem of matching face in VIS spectrum against iris images in NIR spectrum (cross-modal) using periocular information. They utilized LBP, Normalized Gradient Correlation (NGC), and Joint Dictionary-based Sparse Representation (JDSR) methods to accomplish cross-modality matching. 

\item \textbf{Periocular Forensics}: The authors in \citep{Marra2018, Banerjee2018} deduced sensor information from the periocular images. In another work, \cite{Banerjee2019} suppressed the sensor-specific information (sensor anonymization) and also incorporated the sensor pattern of a different device (sensor spoofing) in periocular images.  

\item \textbf{Other Applications}: \cite{Du2016} utilized the periocular region to correct mislabeled left and right iris images in a diverse set of iris datasets. The work in \citep{AlonsoFernandez2014c, Hoffman2019} suggested the use of periocular information for iris spoof detection. \cite{AlonsoFernandez2014c} detected iris spoofs using VIS periocular images, whereas \cite{Hoffman2019} utilized NIR periocular images. \cite{Patel2017} explored the effectiveness of periocular region in verifying kinship using a Block-based Neighborhood Repulsed Metric Learning framework. \cite{JuefeiXu2011b} presented a framework of utilizing the periocular region for age invariant face recognition. The authors applied Walsh-Hadamard transform encoded Local Binary Patterns (WLBP) and Unsupervised Discriminant Projection (UDP), and achieved 100\% rank-1 identification rate on a dataset of 82 subjects. The authors in \citep{Jillela2012, Raja2016c} utilized the periocular region to identify individuals after they undergo facial plastic surgery. \cite{Mahalingam2014} introduced a medically altered gender transformation face dataset and proposed the fusion of periocular (patched-based LBP) with face, which outperformed standalone commercial-off-the-shelf face matchers. \cite{Keshari2016} investigated periocular recognition on pre- and post-cataract surgery images.

\end{enumerate}

\section{Datasets and Competitions}
\label{sec:Periocular-Datasets}

\begin{figure*}[h]
\centering
\includegraphics[width=\textwidth]{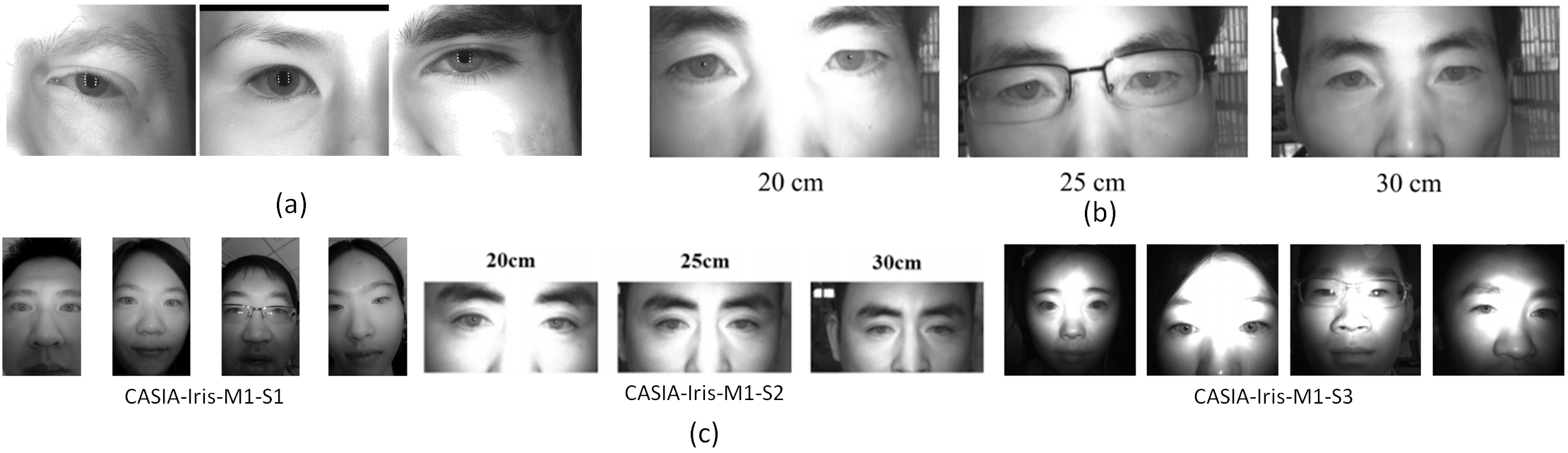}
\caption{Examples of periocular images from NIR datasets: (a) \cite{FOCS} Dataset, (b) MIR 2016 Dataset \citep{Zhang2016}, (c) CASIA-Iris-Mobile-V1.0 Dataset \citep{Zhang2018}.}
\label{fig:periocular-datasets-NIR}
\end{figure*}

\begin{figure*}[h]
\centering
\includegraphics[width=\textwidth]{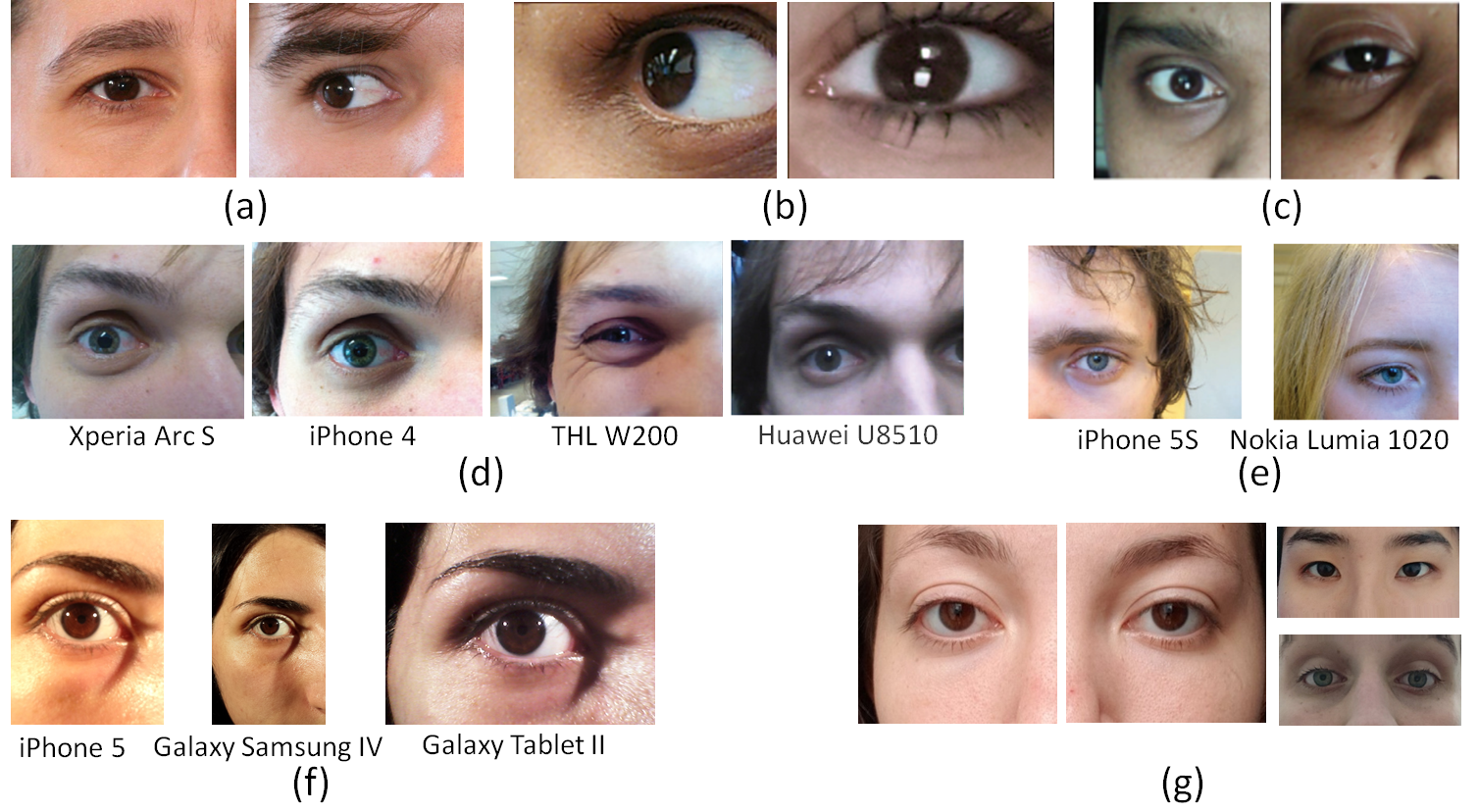}
\caption{Examples of periocular images from VIS datasets: (a) UBIPr dataset \citep{Padole2012}, (b) VISOB 1.0 Dataset \citep{Rattani2016}, (c) \cite{VISOB2} Dataset, (d) CSIP Dataset \citep{Santos2015}, (e) VSSIRIS Dataset \citep{Raja2015d}, (f) MICHE-I Dataset \citep{DeMarsico2015}, (g) UFPR-Periocular \citep{Zanlorensi2020a}.}
\label{fig:periocular-datasets-VIS}
\end{figure*}

\begin{figure*}[h]
\centering
\includegraphics[width=0.8\textwidth]{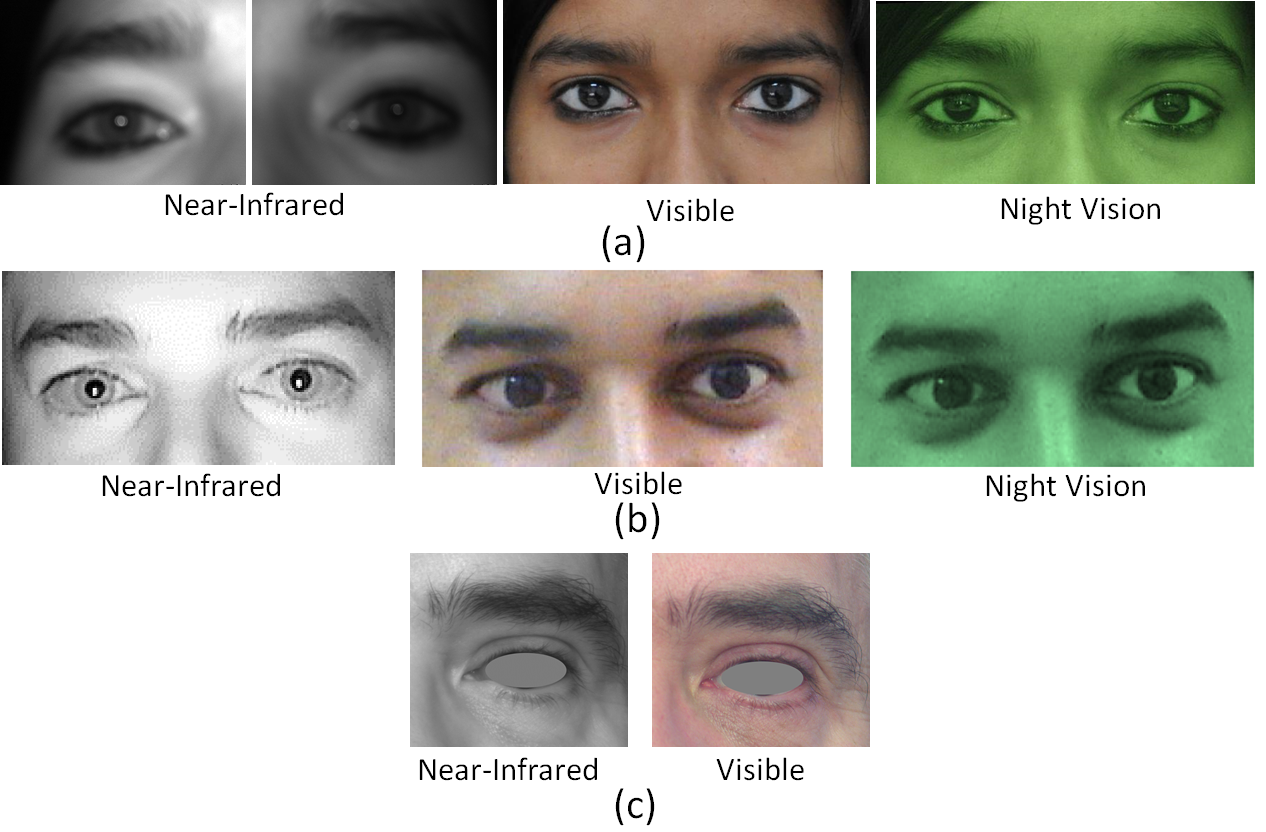}
\caption{Examples of periocular images from Multi-spectral datasets: (a) IIITD Multispectral Periocular (IMP) \citep{Sharma2014}, (b) QUT Multispectral \citep{Algashaam2017a}, and (c) Cross-Eyed 2016 \citep{Sequeira2016}.}
\label{fig:periocular-datasets-MULTI}
\end{figure*}

 In early literature, periocular recognition was performed using face and iris datasets as there were limited datasets available that contained the periocular region only. Commonly used face datasets to perform periocular recognition research on VIS images are FRGC, FERET, FG-NET, MobBIO, and on NIR images are IIT Delhi v1.0, CASIA Interval, BioSec. The iris datasets used for periocular recognition research are UBIRIS v2 (VIS), MBGC (NIR), and PolyU cross-spectral datasets. Table \ref{table:periocular-datasets} describes the datasets specifically collected for periocular recognition. Figures \ref{fig:periocular-datasets-NIR}, \ref{fig:periocular-datasets-VIS}, and \ref{fig:periocular-datasets-MULTI} show a few images from these periocular datasets. The datasets used to perform periocular recognition research under variable stand-off distance are FRGC, UBIRIS v2, and UBIPr. Examples of datasets providing video data of subjects for periocular biometrics research are MBGC, FOCS, and VSSIRIS. Other datasets provide special evaluation scenarios such as aging (MORPH, FG-NET), plastic surgery \citep{Raja2016c}, gender transformation \citep{Mahalingam2014}, expression changes (FaceExpressUBI), face occlusion (AR, Compass), cross-spectral matching (CMU-H, IMP, CROSS-EYED 2016, CROSS-EYED 2017), or mobile authentication (CASIA-Iris-Mobile-V1.0, CSIP, MICHE I and II, VSSIRIS, VISOB 1.0 and 2.0, CMPD). Various competitions focusing on periocular recognition can be found in \citep{Rattani2016,Sequeira2016,DeMarsico2017,Sequeira2017}. The competitions in \citep{Rattani2016, DeMarsico2017} are on mobile periocular images, while the competitions in \citep{Sequeira2016, Sequeira2017} evaluated the cross-spectrum (matching of VIS and NIR images) scenario. Table \ref{table:periocular-competitions} summarizes details about these competitions. \cite{Zanlorensi2021} surveyed various ocular datasets and discussed popular ocular recognition competitions. The authors described 36 iris, 4 iris/periocular, 4 periocular, and 10 multimodal datasets. 

\begin{table*}[]
\caption{Description of periocular datasets (NIR, VIS, and multi-spectrum), along with representative research papers utilizing these datasets.}
\label{table:periocular-datasets}
\resizebox{\textwidth}{!}{%
\begin{tabular}{|l|l|l|}
\hline
\textbf{Datasets} & \textbf{Description} & \textbf{Papers} \\ \hline
\multicolumn{3}{|c|}{\textbf{NIR Spectrum}} \\ \hline
NIST- Face and ocular challenge series \citep{FOCS} & \begin{tabular}[c]{@{}l@{}}9,588 total images, 136 subjects, 750x600 resolution, monocular images\\ IOM sensor\end{tabular} & \begin{tabular}[c]{@{}l@{}}\citep{Boddeti2011,Ross2012,Monwar2013} \\ \citep{Smereka2013,Smereka2015}\\\citep{Verma2016,Zhao2017} \end{tabular}\\ \hline
MIR 2016 \citep{Zhang2016} & \begin{tabular}[c]{@{}l@{}}16,500 total images, 550 subjects, 1968x2014 resolution, biocular images\\ IrisKing mobile sensor\end{tabular} & \citep{Zhang2016} \\ \hline
\begin{tabular}[c]{@{}l@{}}CASIA-Iris-Mobile-V1.0 \\ \citep{Zhang2018}\end{tabular} & \begin{tabular}[c]{@{}l@{}}11,000 total images, 630 subjects, biocular images, CASIA NIR mobile camera\end{tabular} & \citep{Zhang2018} \\ \hline
\cite{CASIAV4} & \begin{tabular}[c]{@{}l@{}}2,567 total images, 142 subjects, 2352x1728 resolution, biocular images, \\ CASIA long-range iris camera\end{tabular} & \citep{Tan2012b,Tan2013,Verma2016} \\ \hline
\multicolumn{3}{|c|}{\textbf{Visible Spectrum}} \\ \hline
UBIPr \citep{Padole2012} & 10,950 total images, 261 subjects, Canon EOS 5D, biocular images & \begin{tabular}[c]{@{}l@{}} \citep{Padole2012,Nie2014,HernandezDiaz2018}\\\citep{Smereka2015,Reddy2019,Reddy2020} \end{tabular}\\ \hline
CSIP \citep{Santos2015} & \begin{tabular}[c]{@{}l@{}}2,004 total images, 50 subjects, monocular images\\ 4 mobile sensors (Xperia Arc S, Apple iPhone4, THL W200, Huawei U8510)\end{tabular} & \citep{Santos2015,Garg2018} \\ \hline
MICHE I \citep{DeMarsico2015} & \begin{tabular}[c]{@{}l@{}}3,732 total images, 92 subjects, monocular images\\ 3 mobile sensors (iPhone5, Galaxy Samsung IV, Galaxy Tablet II)\end{tabular} & \citep{DeMarsico2015,Raja2016d,Reddy2019,Raja2020} \\ \hline
VSSIRIS \citep{Raja2015d} & \begin{tabular}[c]{@{}l@{}}560 total images, 28 subjects, monocular images\\ 2 mobile sensors (iPhone 5S and Nokia Lumia 1020)\end{tabular} & \citep{Raja2015d,AlonsoFernandez2020} \\ \hline
VISOB v1.0\citep{Rattani2016} & \begin{tabular}[c]{@{}l@{}}158,136 total images, 550 subjects, monocular images\\ 3 mobile sensors (iPhone 5s, Samsung Note 4 and Oppo N1)\end{tabular} &  \begin{tabular}[c]{@{}l@{}} \citep{Ahuja2016a,Raghavendra2016a,Ahuja2017} \\ \citep{Garg2018,Hoang2020,Raja2020,Reddy2020} \end{tabular}\\ \hline
CMPD \citep{Keshari2016} & \begin{tabular}[c]{@{}l@{}}2,380 total images, 56 subjects, monocular images\\ MicroMax A350 Canvas Knight mobile device\end{tabular} & \citep{Keshari2016} \\ \hline
MICHE II \citep{DeMarsico2017} & \begin{tabular}[c]{@{}l@{}}3,120 total images, 75 subjects, monocular images, \\ 3 mobile sensors (iPhone5, Galaxy Samsung IV, Galaxy Tablet II)\end{tabular} & \begin{tabular}[c]{@{}l@{}} \citep{Ahuja2016b,Ahmed2017} \\ \citep{Ahuja2017,DeMarsico2017} \end{tabular}\\ \hline
\begin{tabular}[c]{@{}l@{}} UFPR-Periocular \\ \citep{Zanlorensi2020a} \end{tabular}& \begin{tabular}[c]{@{}l@{}}33,660 total images, 1,122 subjects, both monocular and biocular images\\ 196 mobile sensors\end{tabular} & \citep{Zanlorensi2020a} \\ \hline
\cite{VISOB2} & \begin{tabular}[c]{@{}l@{}}75,428 total images, 250 subjects, monocular images\\ 2 mobile sensors (Samsung Note 4 and Oppo N1)\end{tabular} & \citep{Krishnan2020} \\ \hline
\multicolumn{3}{|c|}{\textbf{Multi-spectrum}} \\ \hline
IMP \citep{Sharma2014} & \begin{tabular}[c]{@{}l@{}}1,240 total images, 62 subjects, monocular and biocular images\\ 3 sensors (Cogent iris scanner, Sony HandyCam, Nikon SLR camera)\end{tabular} & \begin{tabular}[c]{@{}l@{}}\citep{Keshari2016,Ramaiah2016,Algashaam2017a}\\\citep{Behera2017,Garg2018,HernandezDiaz2019} \\ \citep{Behera2020,Ipe2020} \end{tabular}\\ \hline
\begin{tabular}[c]{@{}l@{}}CROSS-EYED 2016 \\ \citep{Sequeira2016}\end{tabular} & 3,840 total images, 120 subjects, 900×800 resolution, monocular images & \begin{tabular}[c]{@{}l@{}}\citep{Behera2017,Raja2017,Reddy2018a} \\ \citep{AlonsoFernandez2020,Behera2020,Reddy2020} \end{tabular}\\ \hline
CROSS-EYED 2017 \citep{Sequeira2017} & 5,600 total images, 175 subjects, 900×800 resolution, monocular images & \citep{Sequeira2017} \\ \hline
\begin{tabular}[c]{@{}l@{}} QUT Multispectral \\ \citep{Algashaam2017a} \end{tabular} & \begin{tabular}[c]{@{}l@{}}212 total images, 53 subjects, 800x600 resolution, biocular images\\ 2 sensors (Sony DCR-DVD653E, IP2M-842B Surveillance camera)\end{tabular} & \citep{Algashaam2017a} \\ \hline
\end{tabular}
}
\end{table*}

\begin{table*}[]
\caption{A summary (datasets and performance achieved) of various competitions on periocular recognition. Here, GFRR is Generalized False Rejection Rate, and GFAR is Generalized False Acceptance Rate. }
\label{table:periocular-competitions}
\resizebox{\textwidth}{!}{%
\begin{tabular}{|l|l|l|}
\hline
\textbf{Competition} & \textbf{Dataset} & \textbf{Performance} \\ \hline
MICHE-II \citep{DeMarsico2017} & MICHE-I and MICHE-II & EER is 2.74\% and FRR is 9.13\% @ 0.1\% FAR \citep{Ahmed2016, Ahmed2017} \\ \hline
ICIP \citep{Rattani2016} & VISOB & EER is 0.06\% - 0.20\%  and GMR is 92\%  @ 0.1\% FMR \citep{Raghavendra2016a} \\ \hline
CROSS-EYED 2016 \citep{Sequeira2016} & CROSS-EYED 2016 & GFRR is 0.0\% @ 1\% GFAR and EER is 0.29\% (HH1) \citep{Sequeira2016} \\ \hline
CROSS-EYED 2017 \citep{Sequeira2017} & CROSS-EYED 2016 and 2017 & GFRR is 0.74\% @ 1\% GFAR and EER is 0.82\% (HH1) \citep{Sequeira2017} \\ \hline
\end{tabular}
}
\end{table*}

\section{Challenges and Future Directions}
\label{sec:Periocular-Challenges}
\begin{enumerate}
\item \textbf{Definition and Standardization}: The definition of the periocular region is not standardized. What is the actual boundary around the eye? Should we consider a single eye or both eyes to be in the periocular region? These questions about the scope of the periocular region is yet to be answered. Apart from these definitional concerns, issues around standardization has to be resolved for ground-truth segmentation, and the minimum resolution needed for recognition.

\item \textbf{Generalizability}: Periocular biometric solutions should be generalizable, which refers to the matching of periocular images under cross-sensor (images from different sensors), cross-spectrum (images from different spectra), cross-dataset (images from different datasets), cross-resolution (images at multiple distances), and cross-modal (images from different modalities) scenarios.
    
\item \textbf{Non-ideal Conditions}: Researchers need to focus on periocular matching under non-ideal conditions, i.e., pose variations \citep{Park2011, Karakaya2021}, expression, non-uniform illumination, low-resolution, occlusions (eyeglasses, eye-blinking, different types of masks, scarfs or helmets or eye makeup), or large stand-off distance.
    

\item \textbf{Effects of Aging}: With age, wrinkles and folds around the eye could change the overall appearance of the periocular region. The effects of aging on periocular recognition are yet to be comprehensively studied \citep{Ma2019}. 

    
\item \textbf{Anti-spoofing Measures}: While periocular region has been utilized to detect iris spoof attacks \citep{AlonsoFernandez2014c, Hoffman2019}, we should also be vigilant about spoof attacks directed at the periocular region.

\item \textbf{Explainability and Interpretability}: Increasing use of deep learning-based techniques in periocular biometrics opens another direction which involves explainability of these deep learning models \citep{Brito2021}.
 
\end{enumerate}

\section{Summary}
\label{sec:Periocular-Conclusion}
This article provided a survey on periocular biometrics in the wake of its importance due to the increased use of face masks. Firstly, we reported recent face and periocular recognition techniques specifically designed to recognize humans wearing a face mask. Subsequently, we provided details on various aspects of periocular biometrics, viz., anatomical cues in the periocular region used for recognition, various feature extraction and matching techniques, cross-spectral recognition, its fusion with other biometrics modalities (face or iris), authentication in mobile devices, usefulness of this biometric in other applications, periocular datasets, and competitions. Finally, we discussed the various challenges and future directions to work on. The applicability of the periocular biometrics is likely to extend to other scenarios where only the ocular region of the face may be visible. This could be due to cultural etiquette (e.g., women covering their face) or safety precautions (e.g., surgeons or construction workers covering their nose and mouth).

\bibliographystyle{Periocular_Biometrics}
\bibliography{Periocular_Biometrics}

\end{document}